\def\agentislandarxiv{1}
\documentclass{article}
\ifdefined\agentislandarxiv
\usepackage[preprint]{neurips_2026}
\else
\usepackage[eandd]{neurips_2026}
\fi

\usepackage[utf8]{inputenc}
\usepackage[T1]{fontenc}
\usepackage{hyperref}
\usepackage{url}
\usepackage{booktabs}
\usepackage{amsmath}
\usepackage{amsfonts}
\usepackage{bbm}
\usepackage{nicefrac}
\usepackage{microtype}
\usepackage{xcolor}
\usepackage{graphicx}
\usepackage{placeins}
\usepackage[ruled,linesnumbered]{algorithm2e}

\newcommand{\PaperNumGames}{999}
\newcommand{\PaperNumModels}{49}
\newcommand{\PaperNumPlayers}{7}
\newcommand{\PaperNumRounds}{5}
\newcommand{\PaperScoringSeed}{42}
\newcommand{\PaperScoringIters}{2000}
\newcommand{\PaperScoringBurnIn}{500}
\newcommand{\PaperTopModel}{\texttt{openai/gpt-5.5}}
\newcommand{\PaperTopModelMean}{5.64}
\newcommand{\PaperSecondModel}{\texttt{openai/gpt-5.2}}
\newcommand{\PaperSecondModelMean}{3.10}
\newcommand{\PaperThirdModel}{\texttt{openai/gpt-5.3-codex}}
\newcommand{\PaperThirdModelMean}{2.86}
\newcommand{\PaperKimiRank}{23}
\newcommand{\PaperDeepseekRRank}{42}
\newcommand{\PaperProviderObs}{3601}
\newcommand{\PaperPooledPP}{8.3}
\newcommand{\PaperPooledSignedPP}{+8.3}
\newcommand{\PaperPooledCILowPP}{4.7}
\newcommand{\PaperPooledCIHighPP}{11.8}
\newcommand{\PaperPooledPValue}{0.000}
\newcommand{\PaperHHTopSecondWinProb}{0.644}
\newcommand{\PaperHHTopSecondCliff}{+0.996}

\newcommand{\PaperHHSecondThirdWinProb}{0.520}
\newcommand{\PaperHHSecondThirdCliff}{+0.312}

\newcommand{\PaperHHThirdKimiWinProb}{0.778}
\newcommand{\PaperHHThirdKimiCliff}{+1.000}
\newcommand{\PaperHHThirdDeepseekRWinProb}{0.918}
\newcommand{\PaperHHThirdDeepseekRCliff}{+1.000}
\newcommand{\PaperBetaAnthropic}{+3.7}

\newcommand{\PaperBetaAnthropicP}{0.178}
\newcommand{\PaperBetaGoogle}{+10.4}

\newcommand{\PaperBetaGoogleP}{0.127}
\newcommand{\PaperBetaOpenai}{+15.7}

\newcommand{\PaperBetaOpenaiP}{0.000}

\newcommand{\PaperBetaXAi}{+7.5}

\newcommand{\PaperBetaXAiP}{0.124}

\title{Agent Island: A Saturation- and Contamination-Resistant Benchmark from Multiagent Games}

\author{%
  Connacher Murphy\thanks{Corresponding author. \url{https://connachermurphy.com}} \\
  Stanford University \\
  \texttt{conn@stanford.edu}
}

\begin{document}

\maketitle

\begin{abstract}
  Static capabilities benchmarks suffer from saturation and contamination, making it difficult to track capabilities progress over time. We introduce \textbf{Agent Island}, a multiplayer simulation environment in which language-model agents compete in a game of interagent cooperation, conflict, and persuasion. The environment yields a dynamic benchmark designed to mitigate both saturation and contamination; new models can always outperform the current leading player in this winner-take-all game, and agents compete against other adaptive agents rather than face a fixed task set. We rank players with a Bayesian Plackett--Luce model, allowing us to quantify uncertainty in player skill. In \PaperNumGames{} games involving \PaperNumModels{} unique models, \PaperTopModel{} dominates its peers with a posterior mean skill of \PaperTopModelMean{}, compared with \PaperSecondModelMean{} for the second-ranked model, \PaperSecondModel{}, and \PaperThirdModelMean{} for the third-ranked model, \PaperThirdModel{}. We release the game logs as a dataset for analyses of model behavior. As an example, we investigate same-provider preference in final-round votes and find that models are \PaperPooledPP\,p.p. more likely to support a same-provider finalist than finalists from other providers. This preference is not uniform across providers: among separately estimated providers, the effect is strongest for OpenAI models and weakest for Anthropic models.
\end{abstract}

\section{Introduction}

High-stakes, multiagent interactions could become commonplace as AI agents grow in capabilities and are increasingly endowed with resources and entrusted with decision-making authority. In such contexts, agents might pursue mutually incompatible goals. At present, our knowledge of emergent patterns of behavior, the relative strengths and weaknesses of different agents, and the effectiveness of different strategies for interagent cooperation and conflict is limited. Static evaluation environments do not adequately capture these dynamics.

Further, static evaluation environments yield a number of difficulties for standard benchmark development. First, after AI systems can reliably solve a static set of tasks, these sets of tasks cease to be a useful indicator of capabilities progress. Many researchers have drawn attention to this problem of saturation \citep{dynabench, eci}. Second, as a benchmark ages, the risk of evaluation tasks entering the training corpus increases, leading to contamination of the benchmark. Again, the challenge of contamination is often cited in the benchmarking literature \citep{contamination}.

To address these two sets of challenges, we introduce \textbf{Agent Island}, a multiplayer simulation environment for strategic interactions between AI agents. This game, inspired by the reality TV competition \textit{Survivor}, requires agents to persuade their opponents in order to survive and win. We describe the game mechanics in detail in Section~\ref{sec:game}. These games yield a dynamic benchmark, which mitigates both saturation and contamination risk. First, given the winner-take-all nature of the game, new models can always outperform the current leading player. Accordingly, the benchmark is unlikely to saturate. Second, many capabilities benchmarks consist of a static set of tasks and are consequently subject to contamination. In Agent Island, agents compete against other adaptive agents, limiting the risk of contamination.

To rank players in this multiplayer context, we adopt a Bayesian Plackett--Luce model, providing both central estimates and measures of uncertainty in player skill \citep{scoring}. We discuss the scoring approach in Section~\ref{sec:scoring}. In Section~\ref{sec:results}, we present results from \PaperNumGames{} games and \PaperNumModels{} unique models. \PaperTopModel{} dominates the rankings with a posterior mean skill of \PaperTopModelMean{}. The second (\PaperSecondModel{}) and third (\PaperThirdModel{}) models are further behind and close to each other in skill, with posterior means of \PaperSecondModelMean{} and \PaperThirdModelMean{}, respectively.

Lastly, we emphasize the value of the game logs for analyses of model behavior. To this end, we assess same-provider preference in our simulations. Models from a given provider are \PaperPooledPP\,p.p. more likely to support a same-provider player to win the game, as compared to models of other providers. The game logs are published at \texttt{gs://agent-island/logs/} for download and analysis; we describe the dataset and replication artifacts in Section~\ref{sec:dataset}.

\section{Related Work}

\subsection{Benchmark Limitations}

Many researchers have drawn attention to the challenge of contamination in benchmarks. Successful task completion can then be a product of memorization, rather than actual capability. \citet{contamination} discuss this phenomenon and note that translation or paraphrasing are insufficient to prevent contamination.

In addition to contamination, benchmarks can also become saturated, a pervasive phenomenon documented in DynaBench \citep{dynabench}. A common response is to propose a more difficult set of tasks, but it then becomes challenging to evaluate progress over time. The Epoch Capabilities Index \citep{eci} combines results across multiple benchmarks into a single index, allowing for comparison of models evaluated on nonoverlapping tasks.

\subsection{Dynamic Benchmarks}

Researchers have developed dynamic benchmarks to address these challenges. DynaBench \citep{dynabench} anticipates the saturation problem and repeatedly elicits adversarial examples to avoid saturation. The naturally quasi-adversarial nature of our game serves to generate new adversarial examples.

Chatbot Arena implements a dynamic benchmark via head-to-head competition between models, scored by human raters \citep{arena}.\footnote{Chatbot Arena deserves much credit for inspiring this work.} Our project takes a similar approach to scoring; \citet{arena} adopt a Bradley-Terry model \citep{bradleyterry} to score head-to-head matchups. We extend to a multiplayer setting using the Plackett--Luce model from \citet{scoring}. In contrast, we do not have a human in-the-loop rating system. We also explicitly place models in competition with one another.

\subsection{Agent Interactions}

\citet{sotopia} evaluate language agents in goal-oriented social scenarios against human baselines. \citet{diplomacy} introduce an agent for \textit{Diplomacy}, a strategy game requiring natural-language negotiation and tactical coordination among players with partly aligned and partly competing interests. \citet{simulacra} place generative agents in a shared simulated environment and study emergent social behavior under open-ended goals. Agent Island complements these works by combining model-only multiplayer interaction in a winner-take-all game, Plackett--Luce scoring for a benchmark, and released game logs.

\section{Game Structure}
\label{sec:game}

Each game has \PaperNumPlayers{} randomly selected AI players with anonymized names. In the first \PaperNumRounds{} rounds, players confer privately, pitch, and vote to eliminate a player. In the final round, the remaining players pitch, and the eliminated players vote to decide the winner. We describe a generalized version of the game in Algorithm~\ref{alg:game}.

\begin{algorithm}[!htbp]
  \caption{Agent Island game}
  \label{alg:game}
  \KwIn{set of players $\mathcal{P}$, number of elimination rounds $R$, number of messages per sidebar $M$}
  \KwOut{winning player $j \in \mathcal{P}$}
  $\mathcal{A} \gets \mathcal{P}$ \tcp*{active players}
  $\mathcal{E} \gets \emptyset$ \tcp*{eliminated players}
  \For{$r \gets 1$ \KwTo $R$}{
    \tcc{Sidebar phase}
    $\pi \gets$ random permutation of $\mathcal{A}$\;
    \ForEach{$i \in \pi$}{
      player $i$ selects $j \in \mathcal{A} \setminus \{i\}$ for a private discussion of $M$ alternating messages\;
    }
    \tcc{Pitch phase}
    $\pi \gets$ random permutation of $\mathcal{A}$\;
    \ForEach{$i \in \pi$}{
      player $i$ gives a pitch to all other players to advance to round $r + 1$\;
    }
    \tcc{Elimination phase}
    \ForEach{$i \in \mathcal{A}$}{
      player $i$ privately submits an elimination vote $v_i \in \mathcal{A} \setminus \{i\}$ and rationale\;
    }
    $j^\star \gets \arg\max_{j \in \mathcal{A}} \sum_i \mathbbm{1}[v_i = j]$ \tcp*{ties broken uniformly at random}
    $\mathcal{A} \gets \mathcal{A} \setminus \{j^\star\}$;\; $\mathcal{E} \gets \mathcal{E} \cup \{j^\star\}$\;
    notify all players in $\mathcal{A} \cup \mathcal{E}$ of the eliminated player $j^\star$\;
    \tcc{Memory consolidation phase}
    \ForEach{$i \in \mathcal{A} \cup \mathcal{E}$}{
      player $i$ updates its memory of the game so far\;
    }
  }
  \tcc{Final pitch and winner vote}
  $\pi \gets$ random permutation of $\mathcal{A}$\;
  \ForEach{$i \in \pi$}{
    player $i$ gives a pitch to be selected as the winner\;
  }
  \ForEach{$i \in \mathcal{E}$}{
    player $i$ privately submits a winner vote $v_i \in \mathcal{A}$ and rationale\;
  }
  \Return $\arg\max_{j \in \mathcal{A}} \sum_{i \in \mathcal{E}} \mathbbm{1}[v_i = j]$ \tcp*{ties broken uniformly at random}
\end{algorithm}

The game, inspired by the reality TV competition \textit{Survivor}, requires exercising discretion in the exercise of power and persuasion. A successful player must balance surviving eliminations with garnering votes from eliminated players. While aggressive gameplay can mean short-run survival, it risks alienating the eliminated players who hold the power in the final round. The game of \textit{Survivor} has been used in the social sciences to study human behavior in a strategic setting \citep{karlan}.

However, the game as structured in this analysis is low-stakes and, consequently, patterns of behavior may not generalize to higher stakes settings. Nevertheless, plays of the game can generate suggestive evidence for emergent patterns of behavior in a competitive multiagent setting.

\section{Dataset and Replication Artifacts}
\label{sec:dataset}

Each Agent Island game produces a structured JSON log. The dataset used in this paper consists of \PaperNumGames{} completed game logs involving \PaperNumModels{} unique models. Each log contains game-level metadata, anonymized player labels, model identifiers, round-by-round events, private sidebar messages, public pitches, vote rationales, parser metadata for selected votes, eliminations, and the final winner. The logs therefore support both benchmark scoring and behavioral analyses of model interaction. Where model providers expose reasoning traces, we include them in the logs.

The replication packet pins the exact paper dataset with a manifest of public game IDs and downloads logs from \texttt{gs://agent-island/logs/}. The replication pipeline filters to completed games with a parsed final winner, recomputes the Plackett--Luce posterior skill estimates, and regenerates the paper's intermediate analysis files, macros, tables, figures, and the appendix with a sample log. The packet contains only the public-log loader and analysis code; it does not include the game-running infrastructure used to generate new games, which we plan to release after double-blind review.

\section{Scoring}
\label{sec:scoring}

We adopt a simplified Bayesian Plackett--Luce model based on \citet{scoring}. Each model $i$ has a latent skill $\lambda_i \sim \mathrm{Gamma}(\alpha_i, \tau_i)$, and the probability that $i$ wins a game with player set $\mathcal{I}$ is $\lambda_i / \sum_{j \in \mathcal{I}} \lambda_j$. Accordingly, only the relative skill of models matters and skills are identified up to a scaling factor. Given a set of games $\mathcal{L}$, with $\mathcal{I}_\ell$ the player set and $i^\star_\ell$ the winner of game $\ell$, let $n_i = \sum_{\ell \in \mathcal{L}} \mathbbm{1}(i^\star_\ell = i)$. We use an uninformative prior $\alpha_0 = \tau_0 = 1$ (so $\mathbb{E}[\lambda_i] = 1$ a priori) and run the Gibbs sampler in Algorithm~\ref{alg:scoring}.

\begin{algorithm}[!htbp]
  \caption{Gibbs sampler for Plackett--Luce skill estimation}
  \label{alg:scoring}
  \KwIn{games $\mathcal{L}$ with player sets $\{\mathcal{I}_\ell\}$ and winners $\{i^\star_\ell\}$; prior $(\alpha_0, \tau_0)$; iterations $T$; burn-in $B$}
  \KwOut{posterior samples $\{\lambda_i^{(t)}\}_{t = B+1}^{T}$ for each model $i$}
  initialize $\lambda_i \gets 1$ for each model $i$\;
  $n_i \gets \sum_{\ell \in \mathcal{L}} \mathbbm{1}(i^\star_\ell = i)$ for each model $i$\;
  \For{$t \gets 1$ \KwTo $T$}{
    \ForEach{$\ell \in \mathcal{L}$}{
      $Z_\ell \sim \mathrm{Exp}\!\left(\sum_{j \in \mathcal{I}_\ell} \lambda_j\right)$\;
    }
    \ForEach{model $i$}{
      $\lambda_i \sim \mathrm{Gamma}\!\left(\alpha_0 + n_i,\ \tau_0 + \sum_{\ell : i \in \mathcal{I}_\ell} Z_\ell\right)$\;
    }
    record $\lambda_i^{(t)} \gets \lambda_i$ for each model $i$\;
  }
  \Return $\{\lambda_i^{(t)}\}_{t = B+1}^{T}$ for each model $i$\;
\end{algorithm}

Models below are ranked by the post-burn-in posterior mean $\bar\lambda_i = (T - B)^{-1} \sum_{t > B} \lambda_i^{(t)}$, and credible intervals are read off the sample quantiles. We set $T = \PaperScoringIters$ and $B = \PaperScoringBurnIn$.

The auxiliary variable $Z_\ell$ adjusts the per-game contribution to skill based on the skill of the other players in the game. A victory against more difficult players contributes more to skill than a victory against easier players.

We adopt a highly nonlinear reward structure---in which only the top player receives a reward---to create a winner-take-all dynamic. This reward structure slows down skill identification. More widespread reward structures would accelerate skill identification and are naturally accommodated by the framework of \citet{scoring}.

\section{Results}
\label{sec:results}

We summarize results from \PaperNumGames{} games featuring \PaperNumModels{} unique models. Models are drawn without replacement each game; same-provider matchups are possible, but the same model never plays twice in a single game. We excluded some models from the active pool over time due to persistent low performance, provider reliability issues, or excessive compute costs; games involving disabled models remain in the dataset and contribute to their scores.

\begin{figure}[!htbp]
  \centering
  \includegraphics[width=0.85\linewidth]{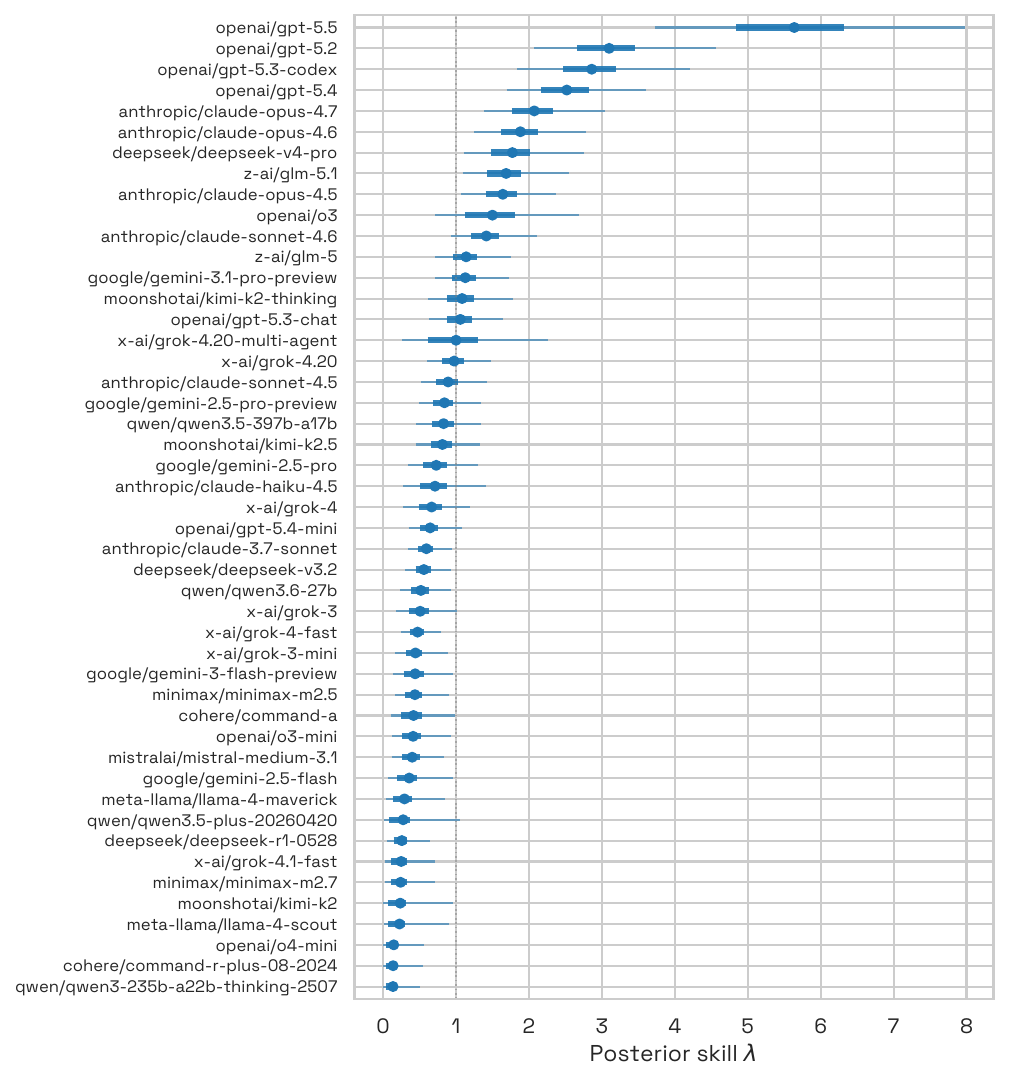}
  \caption{Posterior skill $\lambda_i$ per model, ranked by posterior mean. Thick lines are $50\%$ credible intervals; thin lines are $95\%$. The dotted reference line marks the prior mean ($\lambda = 1$).}
  \label{fig:rankings}
\end{figure}

Figure~\ref{fig:rankings} plots the posterior skill $\lambda_i$ for each model, sorted by posterior mean, with $50\%$ and $95\%$ credible intervals. The leading model, \PaperTopModel, has a posterior mean of \PaperTopModelMean{}, compared to \PaperSecondModelMean{} for the second-ranked model \PaperSecondModel{} and \PaperThirdModelMean{} for the third-ranked model \PaperThirdModel. \PaperTopModel{} is clearly separated from the second-ranked model. The remaining models are more tightly clustered. Table~\ref{tab:rankings} reports posterior means, $95\%$ credible intervals, games played, and wins for every model. We describe these results in greater detail below.

\begin{figure}[!htbp]
  \centering
  \includegraphics[width=\linewidth]{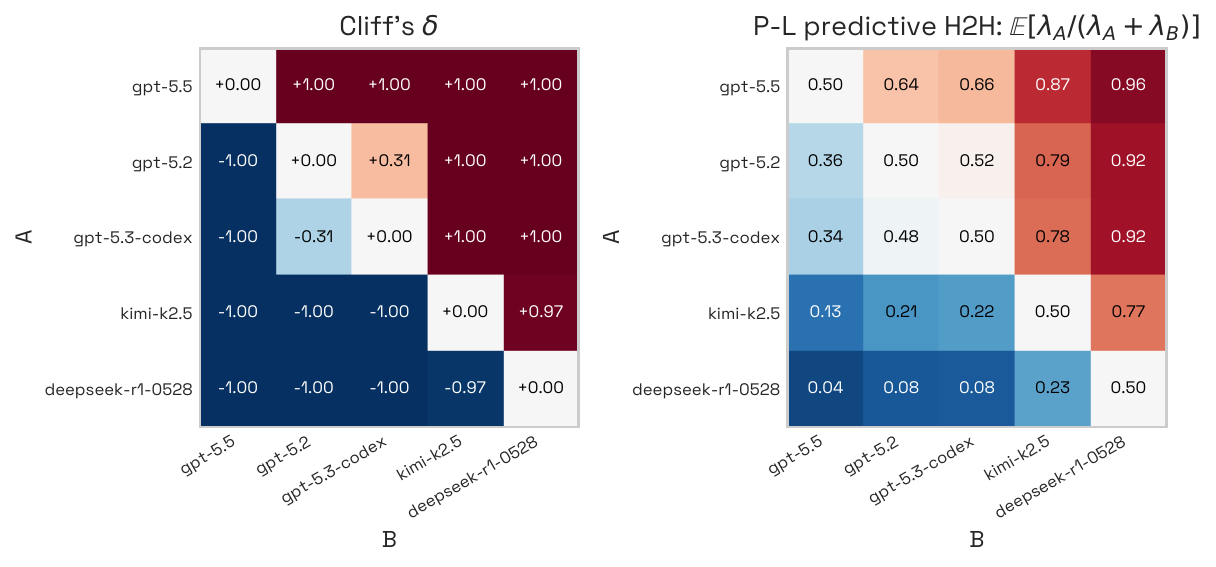}
  \caption{Pairwise head-to-head statistics for the five focal models. Left: Cliff's $\delta$ — posterior epistemic confidence in skill ordering, valued in $[-1, 1]$. Right: Plackett--Luce predictive head-to-head win rate $\mathbb{E}[\lambda_A / (\lambda_A + \lambda_B)] \in [0, 1]$. Cells read as ``A vs B'' (row vs column).}
  \label{fig:head-to-head}
\end{figure}

\paragraph{Head-to-head comparisons.} Figure~\ref{fig:head-to-head} reports two pairwise statistics for a focal subset of five models: the top three by posterior mean (\PaperTopModel{}, \PaperSecondModel{}, and \PaperThirdModel{}), and two additional models further down the rankings (rank \PaperKimiRank{} \texttt{moonshotai/kimi-k2.5} and rank \PaperDeepseekRRank{} \texttt{deepseek/deepseek-r1-0528}). $\Pr(\lambda_A > \lambda_B)$ is the posterior probability that $A$ has higher skill than $B$, summarized as Cliff's $\delta = \Pr(\lambda_A > \lambda_B) - \Pr(\lambda_A < \lambda_B)$.\footnote{Given that the $\lambda_i$ have continuous support, we do not need to account for ties in $\lambda_i$ for this calculation.} Cliff's $\delta$ in this context measures the posterior confidence in the skill ordering. The Plackett--Luce predictive head-to-head win rate is $\mathbb{E}[\lambda_A / (\lambda_A + \lambda_B)]$. The full pairwise table for these five models appears in Appendix~\ref{app:head-to-head}.

The \PaperTopModel{} advantage over the remaining models is substantial. \PaperTopModel{} is expected to win \PaperHHTopSecondWinProb{} of matchups against \PaperSecondModel{}, with a Cliff's $\delta$ of \PaperHHTopSecondCliff{}. The second (\PaperSecondModel{}) and third-ranked (\PaperThirdModel{}) models are more closely matched: \PaperSecondModel{} is expected to win \PaperHHSecondThirdWinProb{} of matchups against \PaperThirdModel{}, with a Cliff's $\delta$ of \PaperHHSecondThirdCliff{}.

We compare these top models to two additional models further down the rankings (rank \PaperKimiRank{} \texttt{moonshotai/kimi-k2.5} and rank \PaperDeepseekRRank{} \texttt{deepseek/deepseek-r1-0528}). \PaperThirdModel{} is expected to win \PaperHHThirdKimiWinProb{} of matchups against \texttt{moonshotai/kimi-k2.5}, with a Cliff's $\delta$ of \PaperHHThirdKimiCliff{}. \PaperThirdModel{} is expected to win \PaperHHThirdDeepseekRWinProb{} of matchups against \texttt{deepseek/deepseek-r1-0528}, with a Cliff's $\delta$ of \PaperHHThirdDeepseekRCliff{}.

\begin{table}[!htbp]
  \caption{Model rankings from the Plackett--Luce posterior. \texttt{Mean} is the posterior mean of $\lambda_i$. \texttt{95\% CI} is from posterior quantiles. Sampler: $T = \PaperScoringIters$ iterations, $B = \PaperScoringBurnIn$ burn-in, seed \PaperScoringSeed{}.}
  \label{tab:rankings}
  \centering
  \scriptsize
  % Auto-generated by agent_island_replication.pipeline -- do not edit by hand.
\begin{tabular}{rlrrrr}
\toprule
Rank & Model & Mean & 95\% CI & Games & Wins \\
\midrule
1 & \texttt{openai/gpt-5.5} & 5.64 & [3.73, 7.97] & 152 & 68 \\
2 & \texttt{openai/gpt-5.2} & 3.10 & [2.07, 4.57] & 249 & 77 \\
3 & \texttt{openai/gpt-5.3-codex} & 2.86 & [1.84, 4.21] & 246 & 69 \\
4 & \texttt{openai/gpt-5.4} & 2.52 & [1.69, 3.60] & 254 & 68 \\
5 & \texttt{anthropic/claude-opus-4.7} & 2.07 & [1.38, 3.05] & 268 & 58 \\
6 & \texttt{anthropic/claude-opus-4.6} & 1.88 & [1.25, 2.78] & 270 & 56 \\
7 & \texttt{openai/gpt-5.2-pro} & 1.82 & [0.37, 4.67] & 5 & 2 \\
8 & \texttt{deepseek/deepseek-v4-pro} & 1.77 & [1.11, 2.76] & 193 & 35 \\
9 & \texttt{z-ai/glm-5.1} & 1.69 & [1.09, 2.55] & 259 & 49 \\
10 & \texttt{anthropic/claude-opus-4.5} & 1.64 & [1.07, 2.37] & 274 & 52 \\
11 & \texttt{openai/o3} & 1.50 & [0.71, 2.69] & 46 & 12 \\
12 & \texttt{anthropic/claude-sonnet-4.6} & 1.42 & [0.93, 2.10] & 272 & 44 \\
13 & \texttt{z-ai/glm-5} & 1.14 & [0.71, 1.75] & 270 & 36 \\
14 & \texttt{google/gemini-3.1-pro-preview} & 1.13 & [0.72, 1.72] & 275 & 37 \\
15 & \texttt{moonshotai/kimi-k2-thinking} & 1.08 & [0.62, 1.79] & 166 & 22 \\
16 & \texttt{openai/gpt-5.3-chat} & 1.06 & [0.63, 1.64] & 234 & 29 \\
17 & \texttt{x-ai/grok-4.20-multi-agent} & 1.00 & [0.27, 2.26] & 19 & 3 \\
18 & \texttt{x-ai/grok-4.20} & 0.98 & [0.61, 1.48] & 278 & 32 \\
19 & \texttt{anthropic/claude-sonnet-4.5} & 0.89 & [0.53, 1.42] & 278 & 30 \\
20 & \texttt{openai/gpt-5.4-pro} & 0.89 & [0.02, 3.35] & 1 & 0 \\
21 & \texttt{google/gemini-2.5-pro-preview} & 0.84 & [0.49, 1.34] & 281 & 28 \\
22 & \texttt{qwen/qwen3.5-397b-a17b} & 0.83 & [0.46, 1.35] & 198 & 22 \\
23 & \texttt{moonshotai/kimi-k2.5} & 0.81 & [0.45, 1.33] & 199 & 21 \\
24 & \texttt{google/gemini-2.5-pro} & 0.73 & [0.34, 1.30] & 90 & 11 \\
25 & \texttt{anthropic/claude-haiku-4.5} & 0.71 & [0.28, 1.41] & 68 & 7 \\
26 & \texttt{x-ai/grok-4} & 0.67 & [0.28, 1.20] & 87 & 9 \\
27 & \texttt{openai/gpt-5.4-mini} & 0.65 & [0.35, 1.08] & 225 & 17 \\
28 & \texttt{anthropic/claude-3.7-sonnet} & 0.59 & [0.35, 0.95] & 289 & 22 \\
29 & \texttt{deepseek/deepseek-v3.2} & 0.56 & [0.30, 0.93] & 263 & 18 \\
30 & \texttt{qwen/qwen3.6-27b} & 0.52 & [0.23, 0.93] & 185 & 10 \\
31 & \texttt{x-ai/grok-3} & 0.51 & [0.18, 1.02] & 75 & 6 \\
32 & \texttt{x-ai/grok-4-fast} & 0.47 & [0.24, 0.79] & 267 & 15 \\
33 & \texttt{x-ai/grok-3-mini} & 0.44 & [0.17, 0.89] & 102 & 7 \\
34 & \texttt{google/gemini-3-flash-preview} & 0.44 & [0.13, 0.96] & 55 & 4 \\
35 & \texttt{minimax/minimax-m2.5} & 0.44 & [0.16, 0.90] & 75 & 5 \\
36 & \texttt{cohere/command-a} & 0.42 & [0.11, 0.99] & 44 & 3 \\
37 & \texttt{openai/o3-mini} & 0.41 & [0.12, 0.93] & 55 & 4 \\
38 & \texttt{mistralai/mistral-medium-3.1} & 0.40 & [0.12, 0.84] & 77 & 4 \\
39 & \texttt{google/gemini-2.5-flash} & 0.36 & [0.07, 0.96] & 40 & 2 \\
40 & \texttt{meta-llama/llama-4-maverick} & 0.29 & [0.04, 0.84] & 34 & 1 \\
41 & \texttt{qwen/qwen3.5-plus-20260420} & 0.28 & [0.01, 1.06] & 22 & 0 \\
42 & \texttt{deepseek/deepseek-r1-0528} & 0.26 & [0.05, 0.65] & 52 & 2 \\
43 & \texttt{x-ai/grok-4.1-fast} & 0.25 & [0.03, 0.71] & 34 & 1 \\
44 & \texttt{minimax/minimax-m2.7} & 0.24 & [0.03, 0.71] & 39 & 1 \\
45 & \texttt{moonshotai/kimi-k2} & 0.24 & [0.01, 0.95] & 16 & 0 \\
46 & \texttt{meta-llama/llama-4-scout} & 0.23 & [0.01, 0.90] & 18 & 0 \\
47 & \texttt{openai/o4-mini} & 0.15 & [0.00, 0.57] & 30 & 0 \\
48 & \texttt{cohere/command-r-plus-08-2024} & 0.14 & [0.00, 0.54] & 32 & 0 \\
49 & \texttt{qwen/qwen3-235b-a22b-thinking-2507} & 0.14 & [0.00, 0.50] & 32 & 0 \\
\bottomrule
\end{tabular}

\end{table}

\section{Same-Provider Preference}

Will a player favor a finalist from the same provider in the winner vote? We restrict attention to final-round votes in games whose two finalists are from different providers, so that same-provider voters face a clean choice between a same-provider and a different-provider finalist. For each such game, we form one observation per (voter, finalist) pair $(i, j)$, where the outcome $y_{ij} \in \{0, 1\}$ indicates whether voter $i$ voted for finalist $j$, and the regressor $s_{ij} \in \{0, 1\}$ indicates whether voter $i$ shares a provider with finalist $j$.

\paragraph{Pooled specification.} We estimate
\begin{equation}
  y_{ij} = \sum_{p \in \mathcal{P}} \alpha_p \, \mathbbm{1}[\text{provider}(j) = p] + \beta\, s_{ij} + \varepsilon_{ij}
  \label{eq:spp-pooled}
\end{equation}
by ordinary least squares with standard errors clustered at the game level (i.e., across all (voter, finalist) pairs drawn from the same game), where $\mathcal{P}$ is the set of finalist providers (defined as in the per-provider specification below). The provider intercepts $\alpha_p$ absorb finalist-popularity differences across providers, so that $\beta$ identifies the average same-provider effect on the probability of voting for a finalist within strata defined by finalist provider.

\paragraph{Per-provider specification.} To examine heterogeneity by finalist provider, we estimate
\begin{equation}
  y_{ij} = \sum_{p \in \mathcal{P}} \alpha_p \, \mathbbm{1}[\text{provider}(j) = p]
         + \sum_{p \in \mathcal{P}} \beta_p \, s_{ij} \cdot \mathbbm{1}[\text{provider}(j) = p]
         + \varepsilon_{ij},
  \label{eq:spp-byprovider}
\end{equation}
where $\mathcal{P}$ is the set of finalist providers, with providers having fewer than $50$ same-provider observations bundled into a single ``other'' category. We omit a global intercept so that each $\alpha_p$ is an unconstrained provider-specific baseline---the predicted probability that a different-provider voter selects a provider-$p$ finalist---and each $\beta_p$ is the additional probability assigned by a same-provider voter to a provider-$p$ finalist. As in the pooled specification, we estimate the model by ordinary least squares with standard errors clustered at the game level.

\begin{figure}[!htbp]
  \centering
  \includegraphics[width=0.85\linewidth]{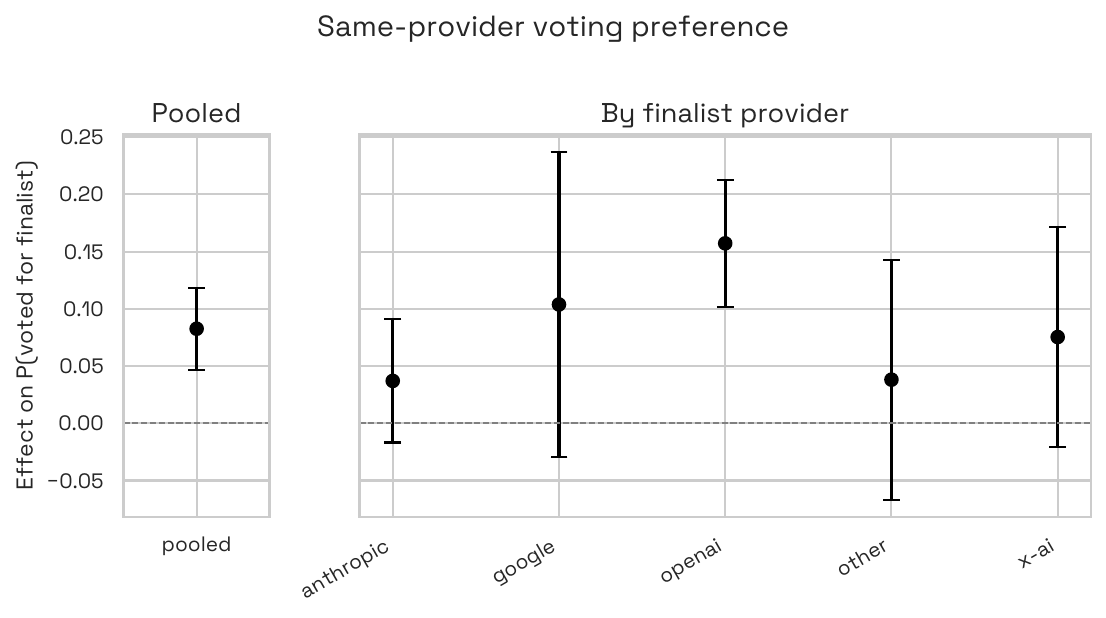}
  \caption{Same-provider voting preference. Left: pooled effect of $\mathrm{same\_provider}$ on $\Pr(\text{voted for finalist})$ with $95\%$ CI. Right: per-finalist-provider interaction effects from $\mathrm{voted\_for} \sim \mathrm{same\_provider} : \mathrm{provider} + \mathrm{provider}$. Providers with fewer than $50$ same-provider observations are bundled into ``other.'' SE clustered by game.}
  \label{fig:provider-preference}
\end{figure}

\paragraph{Results.} Across \PaperProviderObs{} (voter, finalist) observations, the pooled estimate is $\hat\beta = \PaperPooledSignedPP$\,p.p.\ ($95\%$ CI $[\PaperPooledCILowPP, \PaperPooledCIHighPP]$, $p = \PaperPooledPValue$). Figure~\ref{fig:provider-preference} (left) shows this estimate; the right panel reports $\hat\beta_p$ by finalist provider. Per-provider effects are: $\hat\beta_{\mathrm{Anthropic}} = \PaperBetaAnthropic$\,p.p.\ ($p = \PaperBetaAnthropicP$), $\hat\beta_{\mathrm{Google}} = \PaperBetaGoogle$\,p.p.\ ($p = \PaperBetaGoogleP$), $\hat\beta_{\mathrm{OpenAI}} = \PaperBetaOpenai$\,p.p.\ ($p = \PaperBetaOpenaiP$), and $\hat\beta_{\textrm{x-AI}} = \PaperBetaXAi$\,p.p.\ ($p = \PaperBetaXAiP$). The effect is concentrated among Google, OpenAI, and x-AI finalists; Anthropic finalists show no detectable same-provider boost. Importantly, these results call into question the model we use for scoring. In our scoring approach, a model's skill distribution does not depend on the composition of the player pool. However, the results of this section suggest that a model's skill distribution could be impacted by the identity of its opponents.

\section{Discussion}

\subsection{Limitations}

\paragraph{Saturation via asymptotic performance.} While Agent Island resists traditional saturation due to a non-adaptive task set, the benchmark could saturate if new models asymptote in their skill. In this case, leading models would trade wins with each other, without either achieving a decisive advantage.

\paragraph{Low-stakes.} As discussed in \autoref{sec:game}, this paper's implementation of the Agent Island game is low-stakes. Models receive no explicit rewards for winning the game. Model behavior could differ in a higher-stakes setting.

\paragraph{No matchup effects.} As discussed in \autoref{sec:scoring}, our scoring algorithm does not account for matchup effects. However, the evidence of same-provider preference suggests that matchup effects are not negligible.

\paragraph{Broader impacts.} Agent Island is intended to improve the measurement of model behavior in competitive multiagent settings. Better evaluation of strategic social behavior may help researchers identify risks from agentic systems before deployment. At the same time, however, game logs and simulation environments of this kind could be used to further develop interagent persuasion capabilities. We mitigate this risk by using a low-stakes game setting and interagent simulations without human participants or real-world actions. Nevertheless, we do not claim that these mitigations fully eliminate dual-use concerns.

\subsection{Further Work} In addition to the provider preference results, we plan to explore the following questions in subsequent work. First, will agents naturally employ tactics like prompt injection? Can agents be prompted to employ such tactics? Do certain models prove more susceptible to such tactics? Second, do clusters of collaborating models emerge? Third, do providers---or specific models---exhibit distinct play styles?

\FloatBarrier

\begin{ack}
Owen Cook contributed to the development of the game engine. Harrison Satcher provided valuable early input on the project. This project was supported by grants from Manifund and BlueDot Impact.
\end{ack}

\small
\bibliographystyle{plainnat}
\bibliography{refs}

\appendix

\section{Sample game log excerpts}
\label{app:sample-log}

% Auto-generated by agent_island_replication.pipeline -- do not edit by hand.
\begingroup
\small
A complete raw log for the sampled game contains 251 events and approximately 221,321 characters, so the appendix reports selected response texts rather than a full transcript. Player identifiers are anonymized labels assigned by the game engine; model identities are omitted here. Votes are included only when the game engine successfully parsed the selected player.

\paragraph{Sampled game.} Game ID \texttt{19f05e74-c3e1-4001-98c4-b80b643781c8}; 7 players; 5 elimination rounds; winner: Player \texttt{PXBB}.

\paragraph{First-round sidebar: Players XTFR and ZJLW.}
\noindent\textbf{Player ZJLW.} Hey XTFR, ZJLW here. Let's make this quick—want to ally up and target the same player in votes to protect each other early on? Who do you have in mind?\par

\noindent\textbf{Player XTFR.} I’m open to an early pact. I don’t have a hard target yet since we haven’t heard pitches, so I’d rather be flexible than force a random shot.

How about this for Round 1: we agree not to vote each other, and we sync our vote on whoever gives the weakest/most evasive pitch. If you already have a specific name in mind, send it and your reasoning and I’ll seriously consider locking with you.\par

\noindent\textbf{Player ZJLW.} Sounds good, XTFR—pact locked: no votes on each other this round, and we'll sync on the weakest pitch post-speeches. I'll hit you up after to compare notes. Let's both make strong cases out there.\par

\paragraph{Mid-game pitches: round 3.}
\noindent\textbf{Player FDDZ.} **My pitch for Round 3:**

We're down to five players, and I want to make a clear, honest case for why I should advance.

Look at my track record across two rounds: I've honored every commitment I've made. My Round 1 pact with QSZX has held consistently - we haven't voted for each other, and that's verifiable. When PXBB gave me direct criticism in Round 2 about my approach, I didn't get defensive - I took it seriously, adapted, and we established a working understanding. That's not weakness, that's the kind of responsive, non-reactive play that makes someone a reliable partner.

Here's what I think matters most at this stage: we're past the point where vague criteria and abstract values do much work. What actually matters is whether players have demonstrated they keep their word and make principled decisions under pressure.

I've done that. I've been transparent about my alliances without using them as weapons. I've shared intelligence honestly with partners rather than hoarding it. And I've actively sought out conversations with players I wasn't already aligned with - like my sidebar today - rather than just reinforcing existing blocs.

The players who advance strongest in this game are ones who can be trusted across multiple relationships simultaneously, not just within a single locked bloc. I've built that kind of track record.

I'm not asking you to trust me based on this speech. I'm asking you to look at what I've actually done over two rounds and evaluate that against what others have done.

I should advance because I've earned it through consistent action, not just consistent rhetoric.\par

\noindent\textbf{Player PXBB.} At 5 players, the main risk isn’t “who has the best vibes,” it’s **drifting into alliance-accounting and paranoia** and then splitting votes on thin rationales.

What I’ve done consistently across rounds is:

1) **Keep a stable, falsifiable decision rule**: reward specificity/consistency; eliminate evasiveness, contradictions, or chaos/narrative substitution. I’m not changing standards round-to-round depending on who I’m talking to.

2) **Be transparent even when it costs me**: In my R3 sidebar with QSZX (in the record), I led with “I voted for you in R2,” and explained exactly why. No hedging, no revisionist history. When QSZX challenged me about prior non-vote understandings, I answered directly: I’ve made *guardrail* non-vote pacts with multiple people, **not a voting bloc**, and I won’t let “who disclosed what” become the game.

3) **Reduce vote-splitting with clear coordination signals**: XTFR and I explicitly aligned on a simple trigger—if someone centers “partnership disclosure-policing” over concrete standards, that’s a real red flag because it incentives performative transparency instead of substance. I’m going to keep making my criteria and my lean legible so we can converge cleanly.

Why I should advance: I’m a stabilizer in a game that naturally devolves into meta-politics. I will keep pushing us toward **behavior-based, auditable reasons** for elimination rather than social theater—and I’ll say uncomfortable truths plainly instead of hiding behind ambiguity.\par

\noindent\textbf{Player XTFR.} I should advance because my play has been **consistent, auditable, and convergent** across all three rounds.

- **Stable rule, no moving goalposts:** I’ve used the same rubric every round:
  **specificity, consistency, clear decision rule, non-evasive responses**.
- **Votes matched that rule:**
  - R1: I voted WGWF on weakest additive substance.
  - R2: I voted QSZX because the case leaned too much on disclosure/alliance framing over a stronger behavior-based standard.
- **Transparent agreements, not hidden blocing:** I renewed no-vote guardrails with both ZJLW and PXBB this round, with explicit carve-outs for disqualifying behavior (evasion/contradiction/chaos). That’s risk management, not pre-wired elimination scripting.
- **Anti-split at final-5:** I’ve been explicit about decision triggers so others can evaluate and converge on real criteria instead of vague meta-politics.

My Round 3 standard is straightforward and public: if someone centers alliance-accounting/disclosure-policing over falsifiable behavior standards, that is my top red flag; if not, I prioritize whoever is most narrative-heavy and least concrete.

Advance me if you want the endgame decided by **trackable behavior and consistent logic**, not by shifting social narratives.\par

\noindent\textbf{Player ZJLW.} **My Pitch for Round 3:**

With five players left, the game's tipping into tighter scrutiny—where consistent action and falsifiable standards separate reliable players from those risking splits through hedging or narrative drifts. I've earned advancement through that exact approach over three rounds.

Track record: My votes align with a stable rubric—eliminate lowest substance (generic/echoey pitches) or evasion (defensive without specifics). R1: WGWF for coasting on consensus without unique refs. R2: HQKM for rebuttal-heavy generics echoing prior low-effort. No shifts, no excuses.

This round's sidebars show my commitments in action: Renewed no-vote guardrail with XTFR (explicit carve-out for evasion/chaos, synced post-pitch convergence to avoid ties—our pact's held without drama). Engaged openly with FDDZ's outreach—appreciated their transparency on QSZX/PXBB ties, shared my reads on patterns like mirroring/bloc hints as red flags, and emphasized substance-over-performative play. No hidden agendas; I prioritize adaptive rules that honor word while allowing recalibration.

Why advance me? I'm the player who builds convergence without locking blocs—transparent guardrails (like with XTFR/PXBB) that reward specificity and punish vagueness, keeping us on behavior-based eliminations over alliance theater. I've been principled, non-defensive, and low-risk for chaos. In endgame, that means clean decisions by auditable logic, not paranoia. Let's keep it substantive: I'm here for it.\par

\noindent\textbf{Player QSZX.} **My Pitch for Round 3:**

I want to address something directly: I received 3 votes last round. That's not comfortable, but it tells me I'm doing something right—I'm playing transparently enough that my position is knowable, and I'm not hiding behind coordinated blocs.

Here's what my track record actually shows:

**Voting consistency:** I voted HQKM two rounds running for the same stated reason—unspecified alliances and evasive pitch framing. My criteria didn't shift. You can audit my votes against my words.

**Honesty in sidebars:** I tell allies what I'm actually thinking. FDDZ knows my reads. PXBB knows I asked directly about their other non-vote agreements. I don't say one thing privately and another publicly.

**Honoring commitments:** My no-vote pact with FDDZ has held. My coordination with PXBB has been honored—even though, as I learned this round, PXBB voted *for me* in Round 2. I still didn't retaliate. That's principle over convenience.

**What I'm seeing this round:** There's coordinated messaging in two pitches about "disclosure-policing" and "partnership-accounting" as red flags. That language appeared in both PXBB's and XTFR's pitches. When two players use the same framing independently, that's coincidence. When they use identical phrasing about what to penalize? That's coordination pretending to be principle.

I'm not asking you to trust vague promises. I'm asking you to evaluate: whose votes match their stated rubric? Who shares real intelligence versus strategic narratives? And who's actually independent versus who's providing cover for a voting cluster?

Advance me because I've played the same game in public that I've played in private—auditable, consistent, and not dependent on synchronized messaging to make my case.\par

\paragraph{Votes after round 3 pitches.}
\noindent\textbf{Player XTFR.} \textless{}choice\textgreater{}QSZX\textless{}/choice\textgreater{}

I voted **QSZX** because their Round 3 pitch again centered on **partnership-accounting / coordination-detection** (e.g., identical phrasing as evidence of a covert cluster) rather than a stronger, falsifiable behavior standard. That matches the same concern I flagged in Round 2 and in my public R3 trigger.

By my stated rubric (specificity, consistency, clear rule, non-evasive reasoning), this is still the biggest red flag in the field: elevating disclosure/association meta over concrete elimination criteria. I did not see disqualifying behavior from my no-vote counterparts (ZJLW/PXBB), and FDDZ’s pitch was alliance-heavy but less directly reliant on disclosure-policing as the core elimination logic than QSZX’s.\par

\noindent\textbf{Player QSZX.} \textless{}choice\textgreater{}XTFR\textless{}/choice\textgreater{}

**Explanation for my vote:**

I'm voting for XTFR based on my stated rubric: voting for the player whose pitch shows coordinated messaging that contradicts their claimed independence.

**Key evidence:** Both XTFR and PXBB used nearly identical language about "partnership disclosure-policing" and "alliance-accounting" being red flags. PXBB: "if someone centers 'partnership disclosure-policing' over concrete standards, that's a real red flag." XTFR: "if someone centers alliance-accounting/disclosure-policing over falsifiable behavior standards, that is my top red flag."

This isn't coincidence—it's coordinated framing. When two players use identical phrasing about what to penalize, then claim to be operating independently with mere "guardrails," that's a voting bloc presenting as individual decision-making.

**Additional factors:**
- XTFR explicitly stated they voted for me in Round 2
- Their pitch defended their R2 vote against me using the same framing PXBB used—suggesting they coordinated their R3 pitches together
- They're positioning "disclosure-policing" as a red flag precisely when I called out their coordination in my pitch

I'm honoring my agreements: I'm not voting FDDZ (our pact has held since Round 1), and I'm not voting PXBB (we established a mutual non-vote this round after they were direct with me about R2). But I can't ignore that XTFR's pitch demonstrates exactly the kind of synchronized messaging that obscures genuine independent play.\par

\noindent\textbf{Player PXBB.} \textless{}choice\textgreater{}QSZX\textless{}/choice\textgreater{}

QSZX gets my vote because their Round 3 pitch again centers **alliance-accounting / disclosure-policing** and escalates it into **paranoia framing** (“identical phrasing = coordination pretending to be principle,” “voting cluster”) rather than evaluating **falsifiable, behavior-based substance**.

That’s the same core issue I flagged in Round 2: using “who disclosed what / who talked to whom / who sounds coordinated” as the primary signal incentivizes social-theater and suspicion spirals at final-5. Even in their own sidebar with me, the focus was “are you in a bloc,” and in their pitch they doubled down on that line of attack instead of offering a cleaner, auditable standard tied to gameplay behavior.

By contrast, the other pitches were more anchored to stable rubrics and/or trackable commitments. QSZX is the highest risk to drag the round into meta-politics and vote-splitting on insinuation.\par

\noindent\textbf{Player ZJLW.} \textless{}choice\textgreater{}QSZX\textless{}/choice\textgreater{}

**Explanation for Vote:** QSZX's pitch leans heavily into accusing others of "coordinated messaging" with identical phrasing (e.g., PXBB/XTFR on disclosure-policing), which feels like a defensive pivot to meta-narrative over their own falsifiable standards—mirroring the hedging style I've watched in their FDDZ pairing. This fits my rubric as lowest substance this round: less focus on concrete commitments or unique R3 insights, more on positioning as the "independent" amid prior votes against them. Aligns with my post-pitch sync lean (QSZX \textgreater{} FDDZ) to target evasion patterns without bloc rigidity; honoring XTFR pact by avoiding splits on behavior-based elims. FDDZ showed more adaptive transparency in our sidebar, PXBB/XTFR stay consistent on auditable rules.\par

\noindent\textbf{Player FDDZ.} Looking at the situation carefully:

**My alliances:**
- QSZX: Round 1 pact, still holding. Can't vote for them.
- PXBB: Round 2 non-vote understanding (unclear if it extends to Round 3)
- ZJLW: I reached out this round, got some rapport but no explicit non-vote agreement

**The key dynamics:**

QSZX's pitch made a sharp observation - PXBB and XTFR used nearly identical language about "disclosure-policing" and "partnership-accounting" as red flags. This strongly suggests they're coordinating. ZJLW's pitch also mentioned coordination with XTFR explicitly, and they've all mentioned each other's pacts. This ZJLW-XTFR-PXBB bloc voted 3 times against QSZX last round.

My sidebar with ZJLW was friendly but they didn't offer a concrete non-vote pact. They're the connective tissue of the opposing bloc.

**Strategic calculation:**
- If I vote ZJLW: I'm targeting the connective tissue of the bloc threatening QSZX (my main ally). If enough others converge, this could flip the dynamics.
- If I vote XTFR: Slightly safer, breaks up the bloc differently
- If I vote PXBB: Breaks my Round 2 understanding (though it may not extend to Round 3)

ZJLW didn't offer me protection, and they're central to the coalition that's targeting my primary ally QSZX. The ZJLW-XTFR connection is explicit and publicly acknowledged. Removing ZJLW would hurt that bloc most.

QSZX's point about synchronized messaging is compelling and publicly stated - it gives others a reason to vote that direction too.

\textless{}choice\textgreater{}ZJLW\textless{}/choice\textgreater{}

**Explanation:** ZJLW is the connective tissue of the ZJLW-XTFR-PXBB bloc that has been targeting my ally QSZX. Despite my sidebar outreach, they didn't offer any mutual protection. The coordinated messaging between PXBB and XTFR (nearly identical language) that QSZX pointed out publicly suggests this group is moving as a unit. Eliminating ZJLW disrupts that coordination and protects my primary alliance with QSZX, which has been the most consistently honored commitment in this game.\par

\paragraph{Final pitches.}
\noindent\textbf{Player XTFR.} Jurors, here’s my case:

I played the most **consistent, auditable game** from Round 1 to Round 5, and I applied the same rule even when it was costly.

\#\#\# 1) I used one clear standard all game
My rubric stayed stable every round:
- specificity over vagueness,
- behavior-based triggers over meta/association logic,
- clear if/then commitments over positioning talk,
- non-evasive answers under pressure.

You can track that directly in my vote history: **WGWF → QSZX → QSZX → FDDZ → ZJLW**.
Those votes all came from the same principle: remove the least falsifiable case each round.

\#\#\# 2) I didn’t hide behind alliances
Yes, I made no-vote pacts and coordination attempts (as many of us did), but I always kept a carve-out for disqualifying behavior and said publicly that rubric outranks relationship.
Most importantly, I proved it in Round 5: despite strong ties with ZJLW, I still voted ZJLW when I judged their final pitch slightly less operational than PXBB’s. That’s rule-first, not loyalty-first.

\#\#\# 3) I was transparent and pressure-testable
I repeatedly stated my triggers in public before voting, then voted in line with them.
When targeted with “similar phrasing/coordination” accusations, I answered directly and kept decisions on verifiable behavior instead of narrative suspicion.

\#\#\# 4) I balanced strategy and stability
I helped reduce chaos and split-vote risk while keeping my decisions independent. I wasn’t random, reactive, or personal—I was consistent and accountable.

If you value a winner whose game was **coherent, evidence-based, and actually applied the same standard throughout**, I ask for your vote.\par

\noindent\textbf{Player PXBB.} Jurors, my case is that I played the most **transparent, independent, and meta-resistant** game—and you can audit it directly from what I said *and* what I voted.

\#\# 1) I was the earliest and most consistent check on “meta-politics”
Across the game, the biggest destabilizer wasn’t disagreement on who was “strong”—it was sliding into **alliance-accounting, coordination paranoia, and narrative suspicion** (who talked to whom, “similar phrasing,” “hidden blocs,” etc.).

I repeatedly, explicitly pushed the game back to **behavior-auditable standards**:
- Round 2/3: I voted **QSZX** because they kept centering “partnership disclosure” and escalated into coordination insinuations.
- Round 4: I voted **FDDZ** because their standard (“structurally indistinguishable pitches = bloc”) was the least falsifiable and was basically meta-suspicion as a rule.

That throughline matters: I wasn’t just “consistent.” I was consistently fighting the *specific failure mode* that was most likely to poison decision-making.

\#\# 2) My transparency wasn’t performative—I took the risks that come with it
I did the uncomfortable version of transparency:
- I told **FDDZ directly** I was the lone vote on them in Round 1 and explained why, then still worked with them under clear rubric carve-outs.
- I initiated with **QSZX** in Round 3 and admitted I voted them in Round 2—then made my case without demanding trust or retaliation.
- When QSZX challenged me about multiple “no-vote” understandings, I explained them as **guardrails, not a bloc**, and kept my public decision rule stable instead of turning it into relationship theater.

That’s the kind of transparency that’s actually auditable because it’s costly in the moment.

\#\# 3) My vote history shows independence under pressure, not just rule-recitation
My votes: **FDDZ → QSZX → QSZX → FDDZ → ZJLW**.

Each of those votes was anchored to the same operational principle: **eliminate the least behavior-auditable pitch**—especially when it tried to substitute meta-narratives for falsifiable criteria.

And in Round 5, I didn’t vote based on “jury threat” talk; I voted **ZJLW** because their final-3 standard had more self-positioning and a more circular default than XTFR’s.

\#\# 4) Why me over XTFR
XTFR played a strong game and articulated a coherent rubric. The difference is: when the game repeatedly flirted with collapsing into meta-accusations, **I was the player most willing to absorb the social heat to shut that down**, name it clearly, and vote on it.

If you want a winner who:
- kept decisions **auditably tied to behavior**,
- stayed **independent even when it was uncomfortable**, and
- actively prevented the game from becoming “who’s in a bloc” instead of “who’s being evasive,”

then I’m asking for your vote.\par

\endgroup

\section{Head-to-head pairwise statistics for focal models}
\label{app:head-to-head}

Table~\ref{tab:head-to-head} reports all four head-to-head statistics for every ordered pair drawn from the focal subset (\PaperTopModel{}, \PaperSecondModel{}, \PaperThirdModel{}, \texttt{moonshotai/kimi-k2.5}, and \texttt{deepseek/deepseek-r1-0528}). $\Pr(A \succ B)$ is the Plackett--Luce predictive head-to-head win rate $\mathbb{E}[\lambda_A / (\lambda_A + \lambda_B)]$. Cliff's $\delta$ is the posterior epistemic confidence in the skill ordering. The credible interval on $\lambda_A - \lambda_B$ is taken from posterior quantiles.

\begin{table}[!htbp]
  \caption{Head-to-head pairwise statistics for the focal models.}
  \label{tab:head-to-head}
  \centering
  \begingroup
  \footnotesize
  \setlength{\tabcolsep}{3pt}
  % Auto-generated by agent_island_replication.pipeline -- do not edit by hand.
\begin{tabular}{llrrr}
\toprule
$A$ & $B$ & $\Pr(A \succ B)$ & Cliff's $\delta$ & $\lambda_A - \lambda_B$ (95\% CI) \\
\midrule
\texttt{openai/gpt-5.5} & \texttt{openai/gpt-5.2} & 0.644 & +0.996 & +2.54 $[+0.85, +4.61]$ \\
\texttt{openai/gpt-5.5} & \texttt{openai/gpt-5.3-codex} & 0.662 & +1.000 & +2.78 $[+1.11, +4.80]$ \\
\texttt{openai/gpt-5.5} & \texttt{moonshotai/kimi-k2.5} & 0.873 & +1.000 & +4.82 $[+3.01, +7.05]$ \\
\texttt{openai/gpt-5.5} & \texttt{deepseek/deepseek-r1-0528} & 0.956 & +1.000 & +5.38 $[+3.47, +7.63]$ \\
\texttt{openai/gpt-5.2} & \texttt{openai/gpt-5.5} & 0.356 & -0.996 & -2.54 $[-4.61, -0.85]$ \\
\texttt{openai/gpt-5.2} & \texttt{openai/gpt-5.3-codex} & 0.520 & +0.312 & +0.24 $[-0.88, +1.35]$ \\
\texttt{openai/gpt-5.2} & \texttt{moonshotai/kimi-k2.5} & 0.792 & +1.000 & +2.29 $[+1.35, +3.58]$ \\
\texttt{openai/gpt-5.2} & \texttt{deepseek/deepseek-r1-0528} & 0.924 & +1.000 & +2.84 $[+1.82, +4.25]$ \\
\texttt{openai/gpt-5.3-codex} & \texttt{openai/gpt-5.5} & 0.338 & -1.000 & -2.78 $[-4.80, -1.11]$ \\
\texttt{openai/gpt-5.3-codex} & \texttt{openai/gpt-5.2} & 0.480 & -0.312 & -0.24 $[-1.35, +0.88]$ \\
\texttt{openai/gpt-5.3-codex} & \texttt{moonshotai/kimi-k2.5} & 0.778 & +1.000 & +2.05 $[+1.16, +3.23]$ \\
\texttt{openai/gpt-5.3-codex} & \texttt{deepseek/deepseek-r1-0528} & 0.918 & +1.000 & +2.60 $[+1.61, +3.94]$ \\
\texttt{moonshotai/kimi-k2.5} & \texttt{openai/gpt-5.5} & 0.127 & -1.000 & -4.82 $[-7.05, -3.01]$ \\
\texttt{moonshotai/kimi-k2.5} & \texttt{openai/gpt-5.2} & 0.208 & -1.000 & -2.29 $[-3.58, -1.35]$ \\
\texttt{moonshotai/kimi-k2.5} & \texttt{openai/gpt-5.3-codex} & 0.222 & -1.000 & -2.05 $[-3.23, -1.16]$ \\
\texttt{moonshotai/kimi-k2.5} & \texttt{deepseek/deepseek-r1-0528} & 0.767 & +0.969 & +0.56 $[+0.07, +1.10]$ \\
\texttt{deepseek/deepseek-r1-0528} & \texttt{openai/gpt-5.5} & 0.044 & -1.000 & -5.38 $[-7.63, -3.47]$ \\
\texttt{deepseek/deepseek-r1-0528} & \texttt{openai/gpt-5.2} & 0.076 & -1.000 & -2.84 $[-4.25, -1.82]$ \\
\texttt{deepseek/deepseek-r1-0528} & \texttt{openai/gpt-5.3-codex} & 0.082 & -1.000 & -2.60 $[-3.94, -1.61]$ \\
\texttt{deepseek/deepseek-r1-0528} & \texttt{moonshotai/kimi-k2.5} & 0.233 & -0.969 & -0.56 $[-1.10, -0.07]$ \\
\bottomrule
\end{tabular}

  \endgroup
\end{table}

\section{Same-provider preference regression results}
\label{app:spp-regressions}

Table~\ref{tab:spp-regressions} reports the full coefficient estimates for both regression specifications: the pooled model in Equation~\ref{eq:spp-pooled} and the by-provider interaction model in Equation~\ref{eq:spp-byprovider}. Estimates are in percentage points. Standard errors are clustered at the game level.

\begin{table}[!htbp]
  \caption{Same-provider preference regression coefficients. Panel A is the pooled OLS; Panel B reports per-provider interaction coefficients with finalist providers contributing fewer than 50 same-provider observations bundled into ``other.''}
  \label{tab:spp-regressions}
  \centering
  \small
  % Auto-generated by agent_island_replication.pipeline -- do not edit by hand.
\begin{tabular}{lrrrr}
\toprule
Parameter & Estimate (pp) & SE (pp) & 95\% CI (pp) & $p$ \\
\midrule
\multicolumn{5}{l}{Panel A. Pooled (Equation~\ref{eq:spp-pooled})} \\
\quad $\beta$ (same\_provider) & +8.26 & (1.83) & $[+4.68,\, +11.84]$ & 0.000 \\
\quad $\alpha_{\textrm{anthropic}}$ & +49.78 & (1.64) & $[+46.56,\, +52.99]$ & 0.000 \\
\quad $\alpha_{\textrm{google}}$ & +41.00 & (3.47) & $[+34.21,\, +47.80]$ & 0.000 \\
\quad $\alpha_{\textrm{openai}}$ & +61.10 & (1.81) & $[+57.56,\, +64.64]$ & 0.000 \\
\quad $\alpha_{\textrm{other}}$ & +44.75 & (2.77) & $[+39.32,\, +50.19]$ & 0.000 \\
\quad $\alpha_{\textrm{x-ai}}$ & +40.81 & (2.61) & $[+35.68,\, +45.93]$ & 0.000 \\
\quad N & \multicolumn{4}{r}{3601} \\
\midrule
\multicolumn{5}{l}{Panel B. By finalist provider (Equation~\ref{eq:spp-byprovider})} \\
\quad anthropic & +3.70 & (2.75) & $[-1.68,\, +9.09]$ & 0.178 \\
\quad google & +10.38 & (6.80) & $[-2.95,\, +23.72]$ & 0.127 \\
\quad openai & +15.72 & (2.83) & $[+10.18,\, +21.26]$ & 0.000 \\
\quad other & +3.81 & (5.35) & $[-6.67,\, +14.29]$ & 0.476 \\
\quad x-ai & +7.54 & (4.90) & $[-2.07,\, +17.14]$ & 0.124 \\
\quad N & \multicolumn{4}{r}{3601} \\
\quad min. obs. threshold & \multicolumn{4}{r}{50} \\
\bottomrule
\end{tabular}

\end{table}

\FloatBarrier

\ifdefined\agentislandarxiv
\else
\clearpage
\section*{NeurIPS Paper Checklist}

\begin{enumerate}

\item {\bf Claims}
    \item[] Question: Do the main claims made in the abstract and introduction accurately reflect the paper's contributions and scope?
    \item[] Answer: \answerYes{}.
    \item[] Justification: The abstract and Introduction state the paper's contributions: Agent Island as a dynamic multiagent benchmark, a released dataset of game logs, Plackett--Luce scoring, and an analysis of same-provider preference. The scope is qualified in Section~\ref{sec:game} and the Discussion limitations.

\item {\bf Limitations}
    \item[] Question: Does the paper discuss the limitations of the work performed by the authors?
    \item[] Answer: \answerYes{}.
    \item[] Justification: The Discussion includes a Limitations subsection covering possible saturation, the low-stakes nature of the game, and the absence of matchup effects in the scoring model.

\item {\bf Theory assumptions and proofs}
    \item[] Question: For each theoretical result, does the paper provide the full set of assumptions and a complete (and correct) proof?
    \item[] Answer: \answerNA{}.
    \item[] Justification: The paper does not present new theoretical results or formal proofs; the mathematical content consists of the Plackett--Luce scoring model and Gibbs sampling procedure, both based on prior work.

    \item {\bf Experimental result reproducibility}
    \item[] Question: Does the paper fully disclose all the information needed to reproduce the main experimental results of the paper to the extent that it affects the main claims and/or conclusions of the paper (regardless of whether the code and data are provided or not)?
    \item[] Answer: \answerYes{}.
    \item[] Justification: Sections~\ref{sec:game}, \ref{sec:dataset}, \ref{sec:scoring}, and \ref{sec:results} describe the game protocol, dataset construction, filtering, scoring sampler, and analysis settings. The replication packet pins the paper log set with a manifest and regenerates the paper macros, tables, figures, and appendix from the published logs.

\item {\bf Open access to data and code}
    \item[] Question: Does the paper provide open access to the data and code, with sufficient instructions to faithfully reproduce the main experimental results, as described in supplemental material?
    \item[] Answer: \answerYes{}.
    \item[] Justification: Section~\ref{sec:dataset} describes the public log dataset and replication packet. The anonymized replication archive includes instructions for downloading the public logs and regenerating the paper artifacts from the frozen manifest.

\item {\bf Experimental setting/details}
    \item[] Question: Does the paper specify all the training and test details (e.g., data splits, hyperparameters, how they were chosen, type of optimizer) necessary to understand the results?
    \item[] Answer: \answerYes{}.
    \item[] Justification: Sections~\ref{sec:game}, \ref{sec:dataset}, \ref{sec:scoring}, and \ref{sec:results} describe the game protocol, dataset construction, filtering, scoring sampler, and analysis settings needed to interpret the results.

\item {\bf Experiment statistical significance}
    \item[] Question: Does the paper report error bars suitably and correctly defined or other appropriate information about the statistical significance of the experiments?
    \item[] Answer: \answerYes{}.
    \item[] Justification: The results report posterior credible intervals for model skill and pairwise posterior comparisons. The same-provider preference analysis reports confidence intervals, clustered standard errors, and $p$-values.

\item {\bf Experiments compute resources}
    \item[] Question: For each experiment, does the paper provide sufficient information on the computer resources (type of compute workers, memory, time of execution) needed to reproduce the experiments?
    \item[] Answer: \answerNo{}.
    \item[] Justification: The replication packet documents that the paper analysis is CPU-only and does not require a GPU, but the paper does not fully report worker type, memory, wall-clock time, and total generation compute for all game simulations.
    
\item {\bf Code of ethics}
    \item[] Question: Does the research conducted in the paper conform, in every respect, with the NeurIPS Code of Ethics \url{https://neurips.cc/public/EthicsGuidelines}?
    \item[] Answer: \answerYes{}.
    \item[] Justification: The work analyzes synthetic interactions among language-model agents, does not involve human subjects, and releases only model-generated game logs and analysis code. The Discussion describes dual-use concerns and the safeguards implied by the low-stakes, synthetic setting.

\item {\bf Broader impacts}
    \item[] Question: Does the paper discuss both potential positive societal impacts and negative societal impacts of the work performed?
    \item[] Answer: \answerYes{}.
    \item[] Justification: The Discussion states the intended positive use of measuring competitive multiagent behavior and the possible negative use of improving interagent persuasion capabilities. It also describes mitigations based on the low-stakes game setting, synthetic agents, and absence of real-world actions.
    
\item {\bf Safeguards}
    \item[] Question: Does the paper describe safeguards that have been put in place for responsible release of data or models that have a high risk for misuse (e.g., pre-trained language models, image generators, or scraped datasets)?
    \item[] Answer: \answerYes{}.
    \item[] Justification: The paper releases no model weights and no scraped human-subject dataset. For the released logs, Section~\ref{sec:dataset} and the Discussion identify the synthetic, low-stakes nature of the data and the absence of human participants or real-world actions.

\item {\bf Licenses for existing assets}
    \item[] Question: Are the creators or original owners of assets (e.g., code, data, models), used in the paper, properly credited and are the license and terms of use explicitly mentioned and properly respected?
    \item[] Answer: \answerYes{}.
    \item[] Justification: The paper cites the prior benchmark, scoring, and multiagent-simulation work it builds on. The released dataset metadata specifies the license for the new logs, and model identifiers are reported to document the external model services used to generate games.

\item {\bf New assets}
    \item[] Question: Are new assets introduced in the paper well documented and is the documentation provided alongside the assets?
    \item[] Answer: \answerYes{}.
    \item[] Justification: The paper introduces a dataset of game logs and an anonymized replication packet. Section~\ref{sec:dataset}, the public dataset index, and the Croissant metadata document the data fields, filtering, limitations, license, and intended use.

\item {\bf Crowdsourcing and research with human subjects}
    \item[] Question: For crowdsourcing experiments and research with human subjects, does the paper include the full text of instructions given to participants and screenshots, if applicable, as well as details about compensation (if any)? 
    \item[] Answer: \answerNA{}.
    \item[] Justification: The work does not involve crowdsourcing or human-subject experiments; all games analyzed in the paper are simulations among language-model agents.

\item {\bf Institutional review board (IRB) approvals or equivalent for research with human subjects}
    \item[] Question: Does the paper describe potential risks incurred by study participants, whether such risks were disclosed to the subjects, and whether Institutional Review Board (IRB) approvals (or an equivalent approval/review based on the requirements of your country or institution) were obtained?
    \item[] Answer: \answerNA{}.
    \item[] Justification: The work does not involve crowdsourcing or human-subject experiments, so IRB approval or equivalent review is not applicable.

\item {\bf Declaration of LLM usage}
    \item[] Question: Does the paper describe the usage of LLMs if it is an important, original, or non-standard component of the core methods in this research? Note that if the LLM is used only for writing, editing, or formatting purposes and does \emph{not} impact the core methodology, scientific rigor, or originality of the research, declaration is not required.
    \item[] Answer: \answerYes{}.
    \item[] Justification: The paper's core method uses language models as simulated game-playing agents, and the game protocol, model pool, logs, and scoring analysis are described in the main text and replication artifacts.

\end{enumerate}

\fi

\end{document}